%% file: main-arxiv.tex
\documentclass[10pt,twocolumn,letterpaper]{article}

\usepackage{multirow}
\usepackage{subcaption}
\usepackage{mathtools}

\usepackage[pagenumbers]{cvpr} 

\input{preamble}

\definecolor{cvprblue}{rgb}{0.21,0.49,0.74}
\usepackage[pagebackref,breaklinks,colorlinks,allcolors=cvprblue]{hyperref}

\def\paperID{896} 
\def\confName{CVPR}
\def\confYear{2025}

\title{\method: Dynamic Prototype Updating for Multimodal \\ Out-of-Distribution Detection}

\author{
Shawn Li$^1$, Huixian Gong$^1$, Hao Dong$^2$, Tiankai Yang$^1$, Zhengzhong Tu$^3$, Yue Zhao$^1$\\
$^1$University of Southern California \quad $^2$ETH Z\"urich \quad $^3$Texas A\&M University \\
{\tt\small \{li.li02, huixian, tiankaiy, yzhao010\}@usc.edu, hao.dong@ibk.baug.ethz.ch, tzz@tamu.edu}
}

\begin{document}
\maketitle 
\input{./sec/0_abstract_arxiv}
\input{./sec/1_intro}
\input{./sec/2_related_works}
\input{./sec/3_proposed_method}

\input{./sec/4_Experiments}

\input{./sec/5_Conlusion}
{
    \small
    \bibliographystyle{ieeenat_fullname}
    \bibliography{main}
}
\clearpage
\newpage

\appendix
\input{./sec/X_suppl}

\end{document}

%% file: preamble.tex
\usepackage{xspace}
\newcommand{\method}{{DPU}\xspace}

\usepackage{tikz}
\definecolor{darkgreen}{RGB}{60,179,113}
\newcommand{\tikzsquare}[2][red,fill=red]{\tikz[baseline=-0.7ex]\draw[#1] (-#2,-#2) rectangle (#2,#2) ;}%

\definecolor{stepOneColor}{RGB}{140, 185, 215}   
\definecolor{stepTwoColor}{RGB}{225, 175, 160}   
\definecolor{stepThreeColor}{RGB}{225, 205, 130} 

\newtheorem{problem}{Problem}

%% file: sec/0_abstract_arxiv.tex
\begin{abstract}
Out-of-distribution (OOD) detection is crucial for ensuring the robustness of machine learning models by identifying samples that deviate from the training distribution. While traditional OOD detection has predominantly focused on single-modality inputs, such as images, recent advancements in multimodal models have shown the potential of utilizing multiple modalities (e.g., video, optical flow, audio) to improve detection performance. However, existing approaches often neglect intra-class variability within in-distribution (ID) data, assuming that samples of the same class are perfectly cohesive and consistent. This assumption can lead to performance degradation, especially when prediction discrepancies are indiscriminately amplified across all samples. To address this issue, we propose \textbf{Dynamic Prototype Updating} (\method), a novel plug-and-play framework for multimodal OOD detection that accounts for intra-class variations. Our method dynamically updates class center representations for each class by measuring the variance of similar samples within each batch, enabling tailored adjustments. This approach allows us to intensify prediction discrepancies based on the updated class centers, thereby enhancing the model’s robustness and generalization across different modalities. Extensive experiments on two tasks, five datasets, and nine base OOD algorithms demonstrate that DPU significantly improves OOD detection performances, setting a new state-of-the-art in multimodal OOD detection, including improvements up to $80\%$ in Far-OOD detection.
To improve accessibility and reproducibility, our code is released at \url{https://github.com/lili0415/DPU-OOD-Detection}.
\end{abstract}

%% file: sec/1_intro.tex
\vspace{-0.2in}
\section{Introduction}
\vspace{-0.1in}
Out-of-distribution (OOD) detection aims to identify samples that differ from the in-distribution (ID) data in ways that challenge the model's ability to generalize \cite{oodbaseline17iclr,qin2024metaood, grahamdenoising, Li_2023_CVPR,Bai_2024_CVPR}. 
It is crucial for enhancing the safety and robustness of machine learning models across various domains, such as autonomous driving \cite{SuperFusion,li2024light}, 
medical imaging \cite{karimi2022improving},
robotics \cite{blum2019fishyscapes,dong2023jras}, and other applications \cite{cho2023training,yi2023uncertainty,xu2024lego, li2024panoptic,hao2024artificial, Kim_2022_CVPR,Tian_2023_CVPR,Du_2022_CVPR, sun2022icse}.
In recent years, numerous OOD detection algorithms have been developed, spanning classification-based and distance-based methods \cite{yang2022openood,Liang_2018_ECCV,hendrycks2019anomalyseg,energyood20nips,sun2021tone,djurisic2022extremely}. 
Traditionally, OOD detection has focused on \textit{single-modality} inputs, such as images or videos. With the emergence of large Vision-Language Models \cite{clip,NEURIPS2023_407106f4}, researchers are exploring OOD detection with the assistance of language modalities~\cite{ming2022delving,wang2023clipn,Li2024icassp}. However, their evaluations remain limited to benchmarks containing \textit{only images}.
Effectively leveraging multimodal features (\eg, video, optical flow, and audio) remains an open challenge, requiring further research.

\begin{figure}[t]
\centering
\includegraphics[width=0.44\textwidth]{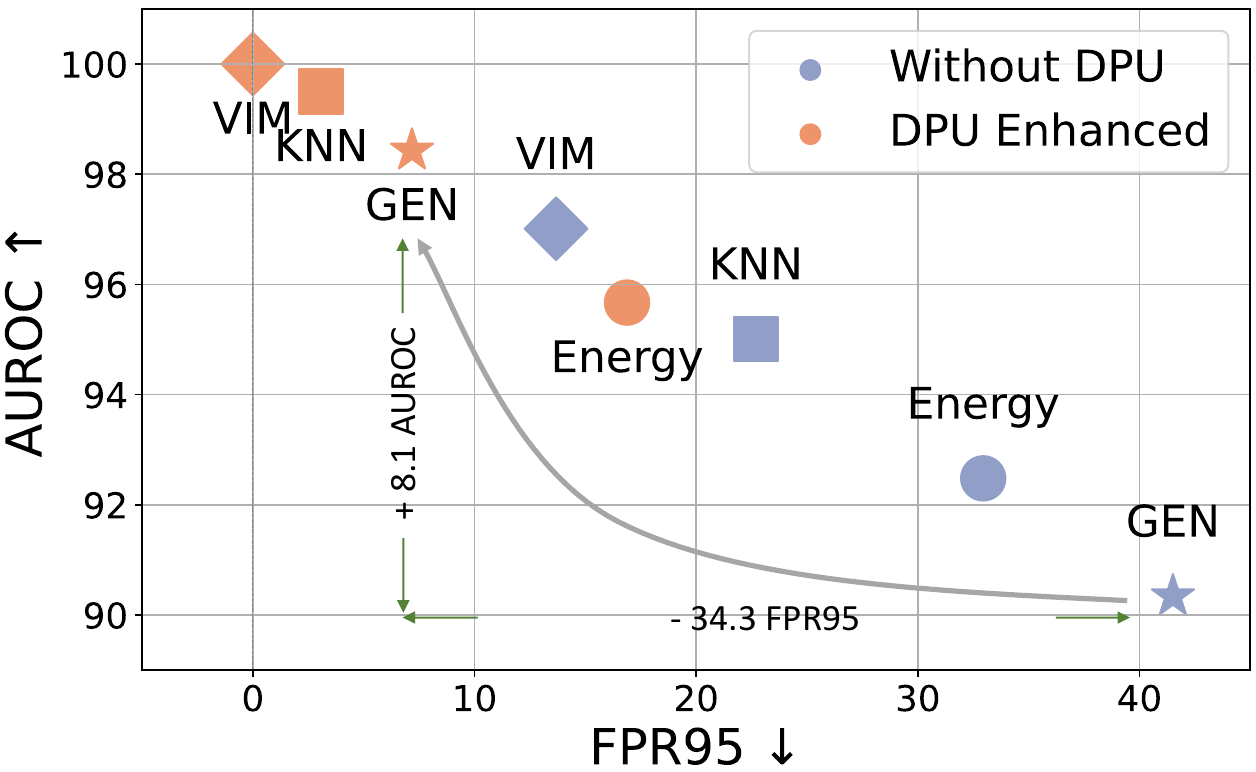} 
\vspace{-0.15in}
   \caption{Performance of our \method applied to four base OOD methods in the Multimodal Far-OOD Detection task (\S \ref{ssec:task-impl-detail}), using HMDB51 as the ID dataset and Kinetics600 as the OOD dataset. Red symbols denote the OOD methods enhanced by DPU, demonstrating that \method significantly improves their performances.
   } 
   \label{Fig:intro1}
   \vspace{-0.15in}
\end{figure}

\noindent
\textbf{Current Work}.
\citet{dong2024multiood} introduced the first multimodal OOD benchmark and framework, identifying the phenomenon of modality prediction discrepancy. This phenomenon reveals that softmax prediction discrepancies across different modalities are negligible for ID data but significant for OOD data. 
By amplifying this prediction discrepancy during training, they observed improvements in OOD detection performance.
However, a key assumption in previous multimodal OOD detection studies is that all samples within a given class are entirely in-distribution \cite{dong2024multiood,yang2022openood}, implying perfect cohesion among samples within the same class. 
This assumption rarely holds true in real-world applications, where intra-class variability is common. 
As a result, applying uniform discrepancy intensification across all training samples can degrade the model's ID prediction accuracy \cite{dong2024multiood}.
When the discrepancy is intensified on class-center samples---typically exhibit consistent predictions across all modalities---this consistency is disrupted, causing significant model confusion and performance degradation, 
as shown in Appx.~\ref{appx:intro-case} case study.

\noindent
\textbf{Our Proposal}.
To tackle the challenge of intra-class variations in existing multimodal OOD detection methods, we introduce a novel approach called \textbf{D}ynamic \textbf{P}rototype \textbf{U}pdating (\method). DPU dynamically adjusts the multimodal prediction discrepancy to ensure high intra-class cohesion and clear inter-class separation, leveraging instance-level training invariance. 
The core idea of \method is to update the prototype representations \cite{li2021adaptive,li2023biased} of each class at a dynamic, sample-specific rate, resulting in more precise and robust model performance. 
These prototype representations act as central reference points, capturing the key features of each class (see Fig. \ref{fig:method} for an overview).
To establish a reliable representation space, we first introduce the Cohesive-Separate Contrastive Training procedure (\S \ref{subsec:contrastive}), which applies marginal contrastive learning to strengthen intra-class cohesion while preserving distinctions between individual samples.
Building on this, we design the Dynamic Prototype Approximation mechanism (\S \ref{subsec:approximation}), which adaptively refines prototype representations based on observed sample variances. This adaptive updating helps mitigate the negative impact of outliers on prototype evolution, stabilizing the learning process.
Using these refined prototypes, we further adjust the multimodal prediction discrepancy for each sample according to its similarity to its class prototype (\S \ref{subsec:proration}). 
Finally, 
OOD models make predictions by leveraging both the joint probability distribution across all modalities and the distinct information from each modality. 
In summary, we make the following contributions:
\begin{itemize}
    \item \textbf{New Observations in Multimodal OOD Detection}. 
    We are the first to identify and explore the negative impact of intra-class variations within ID data for OOD detection.
    \item \textbf{Novel, Model-Agnostic Framework}.
    We propose \method, a flexible, plug-and-play method that effectively handles intra-class variations and is compatible with various existing OOD detection models. As shown in Fig.~\ref{Fig:intro1}, \method enhances performance across various OOD methods.

    \item \textbf{Effectiveness}.
    Comprehensive experiments demonstrate the effectiveness of the proposed method across two tasks, five datasets, and nine base OOD methods. DPU  significantly improves the performance of all benchmark models, achieving new state-of-the-art results, 
    including improvements of around \textbf{10\%} across all metrics for Near-OOD detection and up to \textbf{80\%} for Far-OOD detection.
\end{itemize}

%% file: sec/2_related_works.tex
\section{Related Work (Extended Ver. in Appx.~\ref{appx:related})}
\vspace{-0.05in}
\noindent \textbf{OOD Detection.}
OOD detection identifies samples that deviate  from the training distribution while preserving ID accuracy. Key methods include post hoc techniques such as Maximum Softmax Probability (MSP) \cite{oodbaseline17iclr} and enhancements like temperature scaling \cite{Liang_2018_ECCV}, along with activation-modifying techniques (e.g., ReAct \cite{sun2021tone}) and distance-based approaches (e.g., Mahalanobis \cite{mahananobis18nips}). 
Recent hybrid methods, including VIM \cite{haoqi2022vim}, integrate features and logits, while our method uniquely addresses intra-class variability in ID data to enhance existing base OOD detection methods.


\noindent \textbf{Multimodal OOD Detection.}
Recent efforts extend OOD detection to multimodal contexts, particularly in vision-language systems \cite{ming2022delving,wang2023clipn}. Approaches like Maximum Concept Matching (MCM) \cite{ming2022delving} focus on aligning features to determine OOD scores. 
Additionally, \citet{dong2024multiood} introduced a new multimodal OOD benchmark that includes video, optical flow, and audio, and revealed prediction discrepancies across different modalities.
However, these methods (used as baselines for comparisons in \S \ref{sec:exp}) often overlook the rich contextual information provided by modalities, which our dynamic \method framework leverages.

\vspace{-0.05in}

%% file: sec/3_proposed_method.tex
\section{Proposed \method Framework}
\vspace{-0.05in}

\subsection{Problem Statement and Preliminaries}
\vspace{-0.05in}
Given a training set \( \mathbf{D} = \left\{ (x_i, y_i) \right\}_{i=1}^{n} \), where each \( x_i \in \mathcal{X} \) is an input sample and \( y_i \in \mathcal{Y}=\left\{ 1, 2, \dots, C \right\} \) is a class label, OOD detection aims to distinguish between ID and OOD samples. OOD samples exhibit \textit{semantic shifts} compared to ID samples and do not belong to any class in $\mathcal{Y}$.  The model employs a feature extractor \( g(\cdot) \) to obtain features and a classifier \( h(\cdot) \) to generate predictions, yielding probability output \( \hat{p} \).
When the input data includes multiple modalities, we extend this framework to \textit{multimodal OOD detection}, which we define as follows:
\vspace{-0.1in}

\begin{problem}[Multimodal OOD Detection]
Each training sample \( x_i \) consists of \( M \) modalities, denoted as \( x_i = \left\{ x_{i}^{k} \mid k = 1, \dots, M \right\} \). Multimodal OOD detection combines information from all modalities to make predictions.
\end{problem}
\vspace{-0.05in}

\begin{figure*}[!t]
\begin{center}
\includegraphics[width=\textwidth]{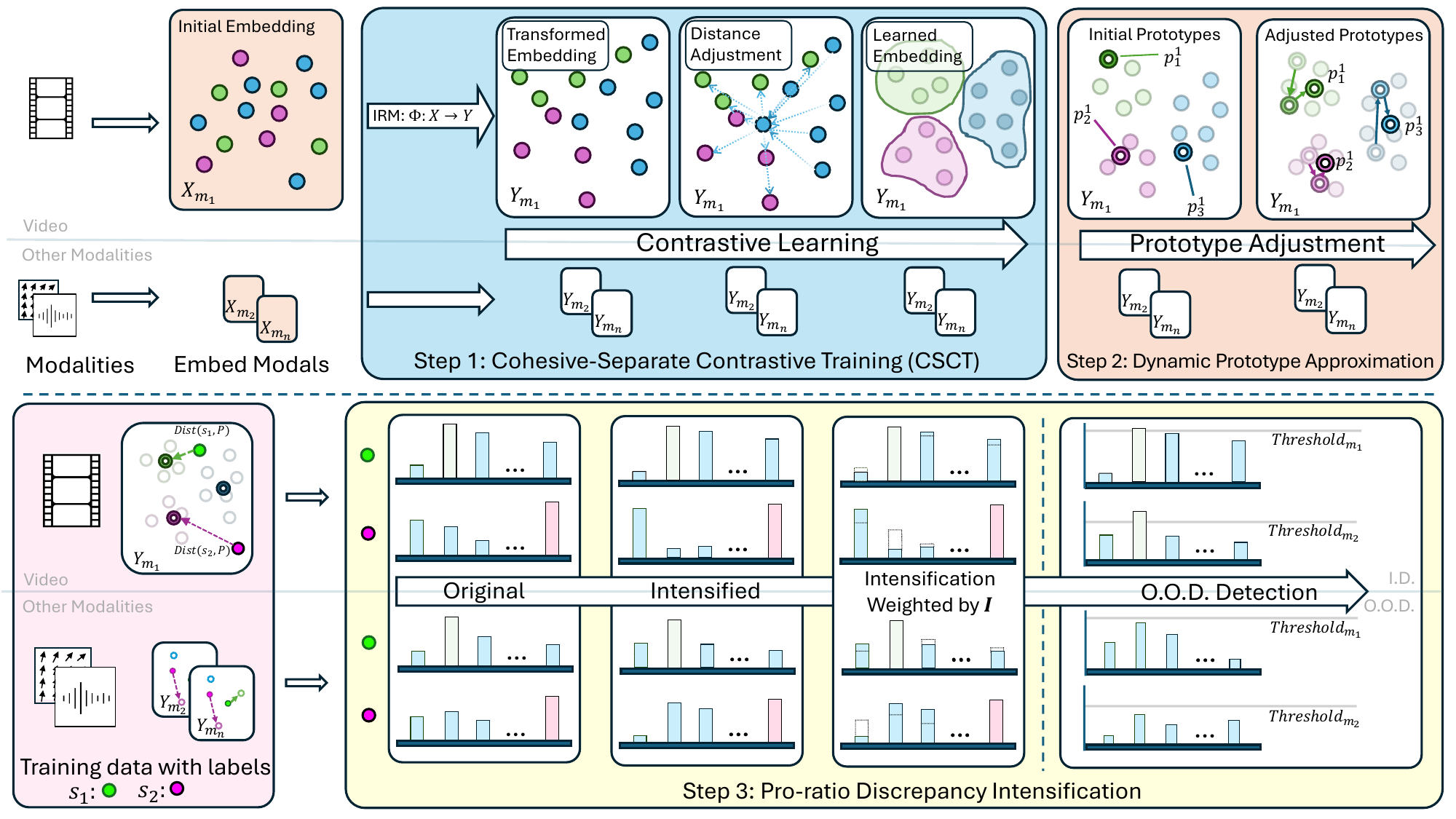} 
   \vspace{-0.3in}
   \caption{ Overview of our \method. 
   It dynamically adjusts multimodal discrepancy intensification based on each sample’s distance to its class prototype. 
   Key components include: (\textcolor{stepOneColor}{Step 1} \tikzsquare[stepOneColor, fill=stepOneColor]{3pt}, {\S \ref{subsec:contrastive}} ) Cohesive-Separate Contrastive Training (CSCT), which aims to preserve intra-class cohesion and inter-class distinctions while capturing within-class variances; 
   (\textcolor{stepTwoColor}{Step 2} \tikzsquare[stepTwoColor, fill=stepTwoColor]{3pt}, {\S \ref{subsec:approximation}}) Dynamic Prototype Approximation (DPA), which refines prototypes to ensure they remain representative despite class outliers; 
   (\textcolor{stepThreeColor}{Step 3} \tikzsquare[stepThreeColor, fill=stepThreeColor]{3pt}, {\S \ref{subsec:proration}}) Pro-ratio Discrepancy Intensification (PDI), which adjusts discrepancy based on sample-prototype similarity, boosting ID accuracy and robustness. Finally, OOD models leverage both joint and modality-specific features for robust OOD detection. 
   }
   \vspace{-0.35in}
   \label{fig:method}
\end{center}
\end{figure*}

The marginal distribution of ID data is denoted as \( P_{\text{in}} \), while OOD samples encountered during testing are drawn from a different marginal distribution \(P_{\text{out}} \). The objective of OOD detection is to construct a decision function \( G \) that classifies each test sample \( x \in \mathcal{X} \) as ID or OOD:
\vspace{-0.1in}

\begin{equation}
    G(x; g, h) = 
        \begin{cases} 
        0 & \text{if } x \sim \mathcal{D}_{\text{out}}, \\
        1 & \text{if } x \sim \mathcal{D}_{\text{in}}.
        \end{cases}
\end{equation}

\noindent
\textbf{Key Challenges in Multimodal OOD Detection}.  
In multimodal OOD detection, previous research has primarily focused on post-processing techniques applied to logit probabilities \cite{oodbaseline17iclr,hendrycks2019anomalyseg,sun2021tone}, with limited emphasis on optimizing the embedding space where learning occurs. While multimodal data offers potential synergy by allowing different modalities to complement each other, how best to leverage this synergy remains an open question.

An essential yet often overlooked challenge is \textit{managing the natural intra-class variations within multimodal ID data}. 
In real-world applications, samples within the same class may exhibit diverse patterns across different modalities, introducing significant variability. The current leading approach, which uniformly amplifies prediction discrepancies across all samples~\cite{dong2024multiood}, risks confusing those samples close to the class center, potentially degrading model performance by weakening class cohesion.
\vspace{-0.05in}

\subsection{Overview of \method}
\vspace{-0.05in}
To address the challenge above, it is key to balance \textit{intra-class cohesion} with \textit{inter-class separation}. We propose a novel approach that dynamically scales the intensification of prediction discrepancies based on each sample’s similarity to its class prototype. 
This adaptive strategy allows samples near the class center to maintain low discrepancy, preserving cohesive predictions, while more distant samples experience higher discrepancy to increase the model’s sensitivity to outlying patterns. This forms the foundation of our proposed \method.
As shown in Fig.~\ref{fig:method}, \method integrates three key components to effectively handle the unique challenges of multimodal OOD detection. The primary objective is to \textit{create a stable representation space with strong intra-class cohesion and distinct inter-class separation}, enabling adaptive discrepancy intensification and enhancing the model’s predictive accuracy across modalities.
\vspace{-0.05in}

As shown in the flowchart, Fig. \ref{fig:method},
the first component, \textit{Cohesive-Separate Contrastive Training (CSCT)} (\S \ref{subsec:contrastive}, \textcolor{stepOneColor}{Step 1} \tikzsquare[stepOneColor, fill=stepOneColor]{3pt} in the figure), enhances intra-class consistency while maintaining clear distinctions between classes, improving the model’s generalization across data distributions. However, class outliers and noise can interfere with prototype representations, reducing reliability. To mitigate this, we introduce \textit{Dynamic Prototype Approximation (DPA)} (\S \ref{subsec:approximation}, \textcolor{stepTwoColor}{Step 2} \tikzsquare[stepTwoColor, fill=stepTwoColor]{3pt} in the figure), which adaptively updates class prototypes by weighting samples according to their similarity to the prototype, ensuring prototypes remain representative of each class.
Additionally, we incorporate \textit{Pro-ratio Discrepancy Intensification (PDI)} and adaptive outlier synthesis (\S \ref{subsec:proration}, \textcolor{stepThreeColor}{Step 3} \tikzsquare[stepThreeColor, fill=stepThreeColor]{3pt} in the figure) to enhance the model’s ability to distinguish ID from OOD samples. These components adjust prediction discrepancies across modalities, strengthening the model’s detection performance. Finally, \method leverages both the joint probability distribution across all modalities and the unique information from each modality to make robust ID/OOD predictions.

\subsection{Step 1: Cohesive-Separate Contrastive Training (CSCT) for Intra-class Cohesion}
\label{subsec:contrastive}

\noindent
\textbf{Motivation.}  
In multimodal OOD detection, it is important to maintain strong intra-class cohesion—ensuring that samples from the same class are represented consistently—while still capturing subtle variations within each class \cite{arjovsky2019invariant, choe2020empirical}. 
Meanwhile, clear separation between different classes is essential for effective differentiation in the learned representation space. Balancing these objectives is key to achieving robust performance across different modalities.

Inspired by \citet{arjovsky2019invariant}, we employ the invariant risk minimization paradigm to (\textit{i}) construct a representation space that is both intra-class cohesive and inter-class separated, and (\textit{ii}) minimize and measure class-wise variances within batches.
Specifically, we define a variant representation function $\Phi : X \to H$ that elicits an invariant predictor $\omega \circ \Phi : H \to Y$ across a set of positive environments $\varepsilon$. This predictor is invariant if the classifier $\omega$ is optimal for all samples within the set. The learning objective is:
\begin{equation}
\begin{aligned}
\min_{\omega , \Phi} \hspace{0.5em}& \sum_{e} R^{e}(\omega \circ \Phi), \nonumber \\
\textrm{s.t.} \hspace{0.5em} &\omega \in \arg\min_{\bar{\omega} : H \to Y} R^{e}(\bar{\omega} \circ \Phi), \hspace{0.5em} \forall e \in \varepsilon.
\end{aligned}
\end{equation}




\noindent Following \cite{arjovsky2019invariant}, this objective can be instantiated as:
\begin{equation}
\label{eq:irmo}
    {\mathcal{L}}_{csct}(\omega, \Phi) = \sum_{e \in \varepsilon} R^e\left( \Phi\right) + \lambda \, \mathbf{D}\left(\omega, \Phi, e\right),
\end{equation}
where $ R^e(\cdot) $ is an empirical-based loss function, $ 
\mathbf{D}(\cdot) $ is a parameterization of invariant risks, and $\lambda$ is a hyperparameter (HP) controlling the balance between objectives. Eq.~(\ref{eq:irmo}) aims to learn a feature representation $\Phi(\cdot)$ that induces a classifier $\omega(\cdot)$, which remains optimal across all samples in the positive environments $e \in \varepsilon$.

To implement Eq.~(\ref{eq:irmo}) in our scenario, we use a two-part strategy. First, we apply robust marginal contrastive learning to construct a representation space that is both cohesive within each class and well-separated between classes. 
This approach aligns naturally with the goals of contrastive learning, which is designed to bring similar samples closer together while pushing dissimilar samples further apart \cite{oord2018representation}.
Second, we measure and minimize class-wise variances within batches to capture subtle intra-class variations and ensure consistency in the learned representations. Together, these steps enable the model to maintain meaningful inter-class distinctions while stabilizing intra-class representations, creating a robust feature space.



\noindent
\textbf{Robust Marginal Contrastive Learning.}  
Consider a batch of video samples \( V_1, V_2, \ldots, V_n \) and their corresponding data from other modalities, \( M_1^k, M_2^k, \ldots, M_n^k \), where \( k \in \{ 1,\dots,M \} \) represents the modality type, and \( n \) is the batch size. For each video sample \( V_j \), we define a positive set \( PS_j \) consisting of samples from the same class and a negative set \( NG_j \) composed of samples from different classes. 

To enhance robustness and sensitivity to inter-class distinctions, we start with the standard InfoNCE loss, a popular contrastive learning objective \cite{oord2018representation}, which is defined as:
\begin{equation}
\label{eq:rmcl}
    {\mathcal{L}}_{rmcl} = \sum_{j=1}^{n} -\log \frac{f_{\text{pos}}}{f_{\text{pos}} + f_{\text{neg}}},
\end{equation}
where \( f_{\text{pos}} \) denotes the similarity scores for samples in the positive set \( PS_j \), and \( f_{\text{neg}} \) represents the similarity scores for samples in the negative set \( NG_j \).

We compute the similarity scores for positive samples using the arc-cosine function, which allows us to effectively measure the angular distance between representations. 
We then add an angular margin \( m \) to further enhance robustness by emphasizing inter-class distinctions. This adjustment helps to accentuate the differences among classes, improving the model's sensitivity to subtle variations within each class.
We defer the details of the modifications to the contrastive loss to Appx.~\ref{appx:robust-margin}.
While contrastive learning can effectively separate different classes, capturing nuanced variations remains challenging, highlighting the need for our refined approach.

\noindent
\textbf{Variance Representation Control.}  
To ensure stable intra-class representations, we introduce an invariant representation regularization term. This regularization encourages invariant representations $\Phi(\cdot)$ to yield consistent prediction distributions across all environments $e \in \varepsilon$. We define it as:
\begin{equation}
\label{eq:irm}
    {\mathcal{L}}_{irm} = \sum_{j=1}^{n} \text{Var}\left(\mathcal{L}^j\right),
\end{equation}
where \( \mathcal{L}^j = \left\{ {\mathcal{L}}_{rmcl}(j) \mid j \in PS_j \right\} \) is the set of loss values for samples within the same positive set. By minimizing the variance of these loss values, the model achieves stable and consistent representation learning for each class \cite{arjovsky2019invariant}.

Combining the objectives from robust marginal contrastive learning and variance representation control, we implement CSCT loss in Eq.~(\ref{eq:irmo}) as:
\begin{equation}
    {\mathcal{L}}_{csct} = \underbracket{{\mathcal{L}}_{rmcl}}_{\text{Eq.~(\ref{eq:rmcl})}} + \lambda \cdot \underbracket{{\mathcal{L}}_{irm}}_{\text{Eq.~(\ref{eq:irm})}},
\end{equation}
where \( \lambda \) balances the influence of variance minimization.

CSCT constructs a representation space that is cohesive within classes and distinct across classes, leveraging robust marginal contrastive learning and variance control. In the next step, we build on this representation space to dynamically update prototype representations for each class.

\subsection{Step 2: Dynamic Prototype Approximation}
\label{subsec:approximation}

\noindent
\textbf{Motivation.}  
Building on the representation space created through CSCT, a dynamic approach to prototype learning is essential for capturing subtle intra-class variations and enhancing class representations. 
By updating prototypes based on sample-wise similarity, we reduce the influence of outliers, allowing each prototype to more accurately represent its class’s central features.

Unlike conventional methods that equally weight all samples within a class, we adjust prototypes according to the variance within each batch, enabling more representative and adaptive prototypes.
Thus, We define the total prototype space as \( P_{\text{ty}_k} \in \mathbf{D}^{L \times Q} \), where \( L \) is the feature dimension, \( Q \) is the number of classes, and \( k \in \{ 1,\dots,M \} \)
denotes the modality. The prototype space is updated dynamically by variance observed during the CSCT process.

For a given batch \( S \), we construct positive sets \( PS = \{ ps_1, ps_2, \dots, ps_P \} \), where each set corresponds to a unique class label, and \( P \) is the number of distinct class labels in \( S \).
For each positive set in \( PS \) with class label \( y \), we calculate the average representation embedding:
\vspace{-0.05in}
\begin{equation}
    H_{\text{av}_k}^{y} = \frac{1}{N^{y}} \sum_{i=1}^{N^{y}} F_i^k, 
\end{equation}
where \( N^{y} \) is the number of samples with label \( y \) in batch \( S \), and \( F_i^k \) represents the embedding (in modality \( k \)) for sample \( i \). This average embedding serves as a central representation for all samples with label \( y \) in the batch.

Using this average embedding, we update the prototype for each class \( y \) through a moving average approach:
\begin{equation}
    P_{\text{ty}_k}^{y} = \beta P_{\text{ty}_k}^{y} + (1 - \beta) \underbracket{\frac{1}{\gamma + \text{Var}(\mathcal{L}^{j}) N^{y}}}_{\text{Update Rate}} \left( H_{\text{av}_k}^{y} - P_{\text{ty}_k}^{y} \right),
\end{equation}
where \( \beta \) and \( \gamma \) are HPs that control the update rate and influence of variance, allowing prototypes to adapt to each batch. 
The update rate is higher when the variance \( \text{Var}(L^{j}) \) is lower or when the batch size \( N^{y} \) is smaller, enabling faster adaptation when the class representations are more consistent or when fewer samples are available.



\subsection{Step 3: Pro-ratio Discrepancy Intensification}
\label{subsec:proration}

\noindent
\textbf{Motivation.}  
In multimodal OOD detection, differences in predictions across various modalities are a strong signal for distinguishing ID samples from OOD samples \cite{dong2024multiood}. The dynamic prototypes we have learned provide a basis for selectively amplifying these prediction discrepancies, enabling the model to more effectively detect OOD samples by emphasizing the separation between ID and OOD data.

To achieve this adaptive intensification, we define an intensification rate  for each sample \( i \) with class label \( y \), which scales the discrepancy by the similarity between the sample's feature representation and its class prototype:
\begin{small}
\begin{equation}
    {\mathcal{L}}_{pdi} = -\underbracket{\mu \cdot \left(1 - \text{Sigmoid}(F_{i}^{v} \cdot (P_{ty_{v}}^{y})^T)\right)}_{\text{Intensification Rate}} \cdot \text{Discr}(\hat{p}_{i}^{k_1}, \hat{p}_{i}^{k_2}),
\end{equation}
\end{small}

\noindent where \( \mu \) is an HP that controls the intensity, $ F_{i}^{v} $ is sample's video feature embedding, \( \text{Sigmoid}(\cdot) \) modulates its similarity to the class prototype, $ \hat{p}_{i}^{k_1} $ and $ \hat{p}_{i}^{k_2} $ denote prediction probabilities of different modalities of sample $ i $, and \( \text{Discr}(\cdot) \) represents a distance metric (we use the Hellinger distance following \cite{dong2024multiood}) that quantifies the discrepancy between probability distributions across modalities. 









\noindent
\textbf{Adaptive Outlier Synthesis.}  
To further improve \method's OOD detection capability, we apply an adaptive outlier synthesis technique. The goal of this technique is to generate synthetic samples that help the model learn to distinguish between ID and OOD data with greater accuracy. Specifically, we create synthetic outliers by fusing prototypes from different modalities and classes.

For each class, we concatenate its prototypes across different modalities, denoted as \( \bar{P} = P_{\text{ty}_{i_1}} \oplus P_{\text{ty}_{i_2}} \), where \( i_1, i_2 \in \{ \text{video, flow, audio} \} \), and \( \oplus \) represents concatenation. 
To form a synthetic outlier, we select a class prototype \( \bar{P}_{y_1} \) and choose another prototype \( \bar{P}_{y_2} \) from the top \( K \) nearest prototypes of \( \bar{P}_{y_1} \). 
We fuse these prototypes as follows:
\begin{equation}
    \bar{P}_{\text{fuse}} = \eta \cdot \bar{P}_{y_1} + (1 - \eta) \cdot \bar{P}_{y_2},
\end{equation}
where \( \eta \) is an HP controlling the fusion balance.

We then divide \( \bar{P}_{\text{fuse}} \) into two separate fused representations, \( \bar{P}_{y_1}^{\text{fuse}} \) and \( \bar{P}_{y_2}^{\text{fuse}} \), and optimize the model using the following objective function:
\begin{small}
\begin{equation}
    {\mathcal{L}}_{aos} = -\left( \text{Discr}(\bar{P}_{y_1}^{\text{fuse}}, \bar{P}_{y_2}^{\text{fuse}}) + E(\bar{P}_{y_1}^{\text{fuse}}) + E(\bar{P}_{y_2}^{\text{fuse}}) \right),
\end{equation}
\end{small}

\noindent where \( E(\cdot) \) is an entropy function that measures prediction uncertainty. By optimizing this objective, the model learns to generalize and distinguish between ID and OOD samples more effectively, leveraging both adaptive intensification and synthetic outlier generation.

\input{./tables/maintable}

Following \cite{dong2024multiood}, our multimodal OOD model leverages the joint probability distribution across all modalities and individual predictions from each modality. The model uses \( M \) modality-specific feature extractors and a shared classifier, with each modality \( k \) also having its own classifier for generating individual prediction probabilities.

The base loss \( \mathcal{L}_{base} \) combines the cross-entropy losses for both joint and individual modality outputs to encourage accurate predictions across modalities.
Finally, the complete objective function for OOD detection integrates additional regularization components from the previous steps:
\begin{equation}
    \mathcal{L} = {\mathcal{L}}_{base} + \delta \cdot {\mathcal{L}}_{csct} + {\mathcal{L}}_{pdi} + \kappa \cdot {\mathcal{L}}_{aos},
\end{equation}
where \( \delta \) and \( \kappa \) are HPs that balance the influence of the CSCT, discrepancy intensification, and adaptive outlier synthesis losses, thereby optimizing the model’s performance on OOD detection.
See details of HP settings in Appx.~\ref{appx:detection}.

%% file: tables/maintable.tex
\begin{table*}[]
\caption{Multimodal Near-OOD Detection results using video and optical flow ($\uparrow$ the higher the better; $\downarrow$ the lower the better). 
`A2D' refers to the Agree-to-Disagree baseline \cite{dong2024multiood}, while `AN' combines Agree-to-Disagree with NP-Mix. \method consistently enhances all metrics across various datasets and OOD detection models, showcasing its effectiveness and adaptability in multimodal OOD detection.} 
\label{TableMain}
\vspace{-0.1in}
\centering
\setlength{\tabcolsep}{4.2pt}
\renewcommand{\arraystretch}{0.95}
\fontsize{7}{10}\selectfont 
\begin{tabular}{r|ccc|ccc|ccc|ccc}
\toprule 
\multicolumn{1}{c|}{\multirow{2}{*}{Method}} & \multicolumn{3}{c|}{HMDB51 25/26}                & \multicolumn{3}{c|}{UCF101 50/51}                & \multicolumn{3}{c|}{Kinetics600 129/110}         & \multicolumn{3}{c}{EPIC-Kitchen} \\ \cline{2-13} 
\multicolumn{1}{c|}{}                        & FPR95 $\downarrow$         & AUROC $\uparrow$         & ID ACC $\uparrow$        & FPR95 $\downarrow$         & AUROC $\uparrow$         & ID ACC $\uparrow$        & FPR95 $\downarrow$         & AUROC $\uparrow$         & ID ACC $\uparrow$        & FPR95 $\downarrow$    & AUROC $\uparrow$   & ID ACC $\uparrow$   \\ \midrule
\multicolumn{1}{l|}{MSP \cite{oodbaseline17iclr}}                     & 44.66          & 87.74          & 89.32          & 22.14          & 95.73          & 99.22          & 64.08          & 76.06          & 80.11          & 76.31      & 67.59     & 71.46      \\
+A2D                                         & 38.78          & 88.37          & 90.64          & 7.09           & 98.19          & 99.61          & 63.04          & 76.47          & 79.50          & 66.23      & 67.91     & 71.83      \\
+AN                                          & 33.99          & 88.79          & 89.98          & 7.96           & 98.24          & 99.71          & 62.91          & 76.92          & 80.52          & 67.91      & 71.52     & 71.64      \\
+Ours                                        & 34.20          & \textbf{89.15} & \textbf{92.16} & \textbf{7.57}  & 98.17          & \textbf{99.81} & \textbf{61.59} & \textbf{77.50} & \textbf{81.07} & \textbf{63.81}      & \textbf{71.47}     & \textbf{72.39}      \\ \midrule
\multicolumn{1}{l|}{Energy \cite{energyood20nips}}                  & 43.36          & 87.46          & 89.32          & 22.52          & 96.06          & 99.22          & 68.75          & 75.49          & 80.11          & 76.68 & 68.29 & 71.46      \\
+A2D                                         & 39.22          & 88.84          & 90.63          & 9.81           & 98.16          & 99.61          & 64.59          & 76.45          & 79.50          &    66.98      & 72.45     & 71.46   \\
+AN                                          & 36.38          & 88.91          & 89.98          & 6.50           & 98.48          & 99.71          & 63.69          & 77.11          & 80.52          &    67.91 & 73.79 & 71.64   \\
+Ours                                        & \textbf{35.07} & \textbf{89.52} & \textbf{92.16} & 7.48           & \textbf{98.43} & \textbf{99.81} & \textbf{62.56} & \textbf{77.40} & \textbf{81.07} & \textbf{63.62}      & \textbf{74.13}     & \textbf{72.39}      \\ \midrule
\multicolumn{1}{l|}{Maxlogit \cite{hendrycks2019anomalyseg}}                & 43.36          & 87.75          & 89.32          & 22.52          & 96.02          & 99.22          & 68.73          & 75.98          & 80.11          & 76.68      & 68.29     & 71.46      \\
+A2D                                         & 39.22          & 88.93          & 90.63          & 9.81           & 98.15          & 99.61          & 64.57          & 76.92          & 79.50          & 66.98      & 72.23     & 71.46      \\
+AN                                          & 36.38          & 89.06          & 89.98          & 6.50           & 98.49          & 99.71          & 63.65          & 77.55          & 80.52          & 66.98      & 73.48     & 71.64      \\
+Ours                                        & \textbf{35.08} & \textbf{89.60} & \textbf{92.16} & 7.57           & 98.39          & \textbf{99.81} & \textbf{62.46} & \textbf{77.76} & \textbf{81.07} & \textbf{63.62}      & \textbf{73.50}     & \textbf{72.39}      \\ \midrule
\multicolumn{1}{l|}{Mahalanobis \cite{mahananobis18nips}}             & 40.31          & 85.28          & 89.32          & 12.14          & 97.14          & 99.22          & 93.51          & 35.83          & 80.11          &  98.69 & 42.99 & 71.46      \\
+A2D                                         & 44.88          & 86.99          & 90.63          & 8.74           & 98.07          & 99.61          & 92.86          & 50.65          & 79.50          & 95.52 &	44.43 &	71.46       \\
+AN                                          & 41.61          & 87.69          & 89.98          & 7.86           & 97.99          & 99.71          & 92.49          & 50.05          & 80.52          & 94.78 &	44.37 &	71.64      \\
+Ours                                        & \textbf{36.17} & \textbf{89.53} & \textbf{92.16} & \textbf{6.60}  & \textbf{98.68} & \textbf{99.81} & \textbf{89.27} & \textbf{54.17} & \textbf{81.07} & 95.34      & \textbf{45.62}     & \textbf{72.39}      \\ \midrule
\multicolumn{1}{l|}{ReAct \cite{sun2021tone}}                   & 42.05          & 87.79          & 89.32          & 25.63          & 95.85          & 99.32          & 72.40          & 73.80          & 80.35          &  83.96 & 65.89 & 71.08      \\
+A2D                                         & 37.91          & 89.09          & 90.63          & 10.39          & 98.12          & 99.61          & 71.26          & 74.29          & 79.64          & 68.66 &	72.03 &	70.52       \\
+AN                                          & 37.47          & 88.63          & 90.20          & 12.43          & 97.33          & 99.90          & 67.95          & 75.55          & 80.70          & 66.60 &	73.11 &	72.39       \\
+Ours                                        & \textbf{35.08} & \textbf{89.51} & \textbf{92.16} & \textbf{7.48}  & \textbf{98.43} & \textbf{99.81} & \textbf{62.57} & \textbf{77.41} & \textbf{81.07} & 71.46      & 71.24     & 69.78      \\ \midrule
\multicolumn{1}{l|}{ASH \cite{djurisic2022extremely}}                     & 53.59          & 87.16          & 89.54          & 32.14          & 94.02          & 99.22          & 69.24          & 76.16          & 79.62          &  76.87 & 67.92 & 70.15     \\
+A2D                                         & 42.05          & 87.72          & 90.41          & 12.52          & 97.43          & 99.42          & 64.28          & 77.28          & 79.15          & 63.25 &	74.73 &	67.72      \\
+AN                                          & 36.17          & 89.30          & 89.32          & 10.68          & 97.80          & 99.81          & 63.77          & 77.44          & 79.44          & 66.23 &	73.06 &	66.23       \\
+Ous                                         & \textbf{35.08} & \textbf{89.51} & \textbf{92.16} & \textbf{7.48}  & \textbf{98.43} & \textbf{99.81} & \textbf{62.57} & \textbf{77.41} & \textbf{81.07} & 71.46      & 71.24     & 69.78      \\ \midrule
\multicolumn{1}{l|}{GEN \cite{liu2023gen}}                     & 43.79          & 87.49          & 89.32          & 23.79          & 95.54          & 99.22          & 69.03          & 75.33          & 80.11          &  76.87 & 68.52 & 71.46      \\
+A2D                                         & 38.56          & 88.61          & 90.63          & 8.83           & 98.12          & 99.61          & 62.28          & 77.08          & 79.50          & 65.86 &	73.05 &	71.46       \\
+AN                                          & 35.95          & 89.78          & 89.32          & 7.67           & 92.28          & 99.71          & 62.95          & 76.95          & 80.52          & 65.30 &	75.17 &	71.64       \\
+Ours                                        & \textbf{34.64} & 89.71          & \textbf{92.16} & \textbf{6.41}  & \textbf{98.41} & \textbf{99.81} & \textbf{62.10} & \textbf{78.05} & \textbf{81.07} & \textbf{63.06}      & 74.22     & \textbf{72.39}      \\ \midrule
\multicolumn{1}{l|}{KNN \cite{sun2022knnood}}                     & 42.92          & 88.46          & 89.32          & 15.63          & 96.93          & 99.22          & 68.67          & 74.64          & 80.11          & 75.93 & 63.60 & 71.46      \\
+A2D                                         & 33.33          & 89.59          & 90.63          & 7.48           & 98.39          & 99.61          & 65.71          & 76.23          & 79.50          & 72.39 &	67.83 &	71.46       \\
+AN                                          & 33.77          & 90.05          & 89.98          & 12.04          & 97.65          & 99.71          & 66.81          & 74.19          & 80.52          & 71.27 &	69.94 &	71.64       \\
+Ours                                        & \textbf{32.90} & 89.36          & \textbf{92.16} & \textbf{11.84} & \textbf{97.74} & \textbf{99.81} & \textbf{66.71} & \textbf{75.34} & \textbf{81.07} & 71.64      & 68.03     & \textbf{72.39}      \\ \midrule
\multicolumn{1}{l|}{VIM \cite{wang2022ofa}}                     & 36.82          & 88.06          & 89.32          & 12.52          & 97.66          & 99.22          & 68.77          & 75.47          & 80.11          & 77.05 & 65.60 & 71.46      \\
+A2D                                         & 33.77          & 89.37          & 90.63          & 6.80           & 98.62          & 99.61          & 64.59          & 76.45          & 79.50          & 66.42 &	68.43 &	71.46       \\
+AN                                          & 34.64          & 88.80          & 89.98          & 4.56           & 98.73          & 99.71          & 63.67          & 77.12          & 80.52          & 67.72 &	69.09 &	71.64       \\
+Ours                                        & \textbf{32.46} & \textbf{89.93} & \textbf{92.16} & 6.79           & 98.56          & \textbf{99.81} & \textbf{62.32} & \textbf{77.51} & \textbf{81.07} & \textbf{66.60}      & \textbf{70.50}     & \textbf{72.39}     \\ \bottomrule 
\end{tabular}
\vspace{-0.15in}
\end{table*}

%% file: sec/4_Experiments.tex
\section{Experimental Results}
\label{sec:exp}
\vspace{-0.05in}
\subsection{Datasets and Evaluation Metrics}
\label{subsec:data-eva}
\vspace{-0.05in}

\noindent \textbf{Datasets.}
Following \cite{dong2024multiood}, we evaluate our method on 5 datasets: HMDB51 \cite{kuehne2011hmdb}, UCF101 \cite{soomro2012ucf101}, Kinetics-600 \cite{kay2017kinetics}, HAC \cite{dong2023SimMMDG}, and EPIC-Kitchens \cite{Damen2018EPICKITCHENS}.  See details in Appx.~\ref{appx:datasets}.

\input{./tables/table2}

\noindent \textbf{Evaluation Metrics.}
We evaluate the performance via the use of the following metrics: (1) the false positive rate (FPR95, the lower, the better) of OOD samples when the true positive rate of ID samples is at 95\%, (2) the area under the receiver operating characteristic curve (AUROC), and (3) ID classification accuracy (ID ACC). 

\subsection{Tasks and Implementation Details}
\label{ssec:task-impl-detail}
\noindent \textbf{Tasks.}
We evaluate \method on two tasks: Near-OOD detection and Far-OOD detection \cite{dong2024multiood}. We present the results on \{video, optical flow\} in the main paper, and defer the results on full modality including \{video, optical flow, audio\} to Appendix. See details in Appx.~\ref{appx:tasks}.

\noindent \textbf{Base OOD Methods and Baselines.}
We include MSP \cite{oodbaseline17iclr}, Energy \cite{energyood20nips}, Maxlogit \cite{hendrycks2019anomalyseg}, Mahalanobis \cite{mahananobis18nips}, ReAct \cite{sun2021tone}, ASH \cite{djurisic2022extremely}, GEN \cite{liu2023gen}, KNN \cite{sun2022knnood}, and VIM \cite{wang2022ofa} as our base OOD methods to be extended to the multimodal setting. 
The training algorithms range from probability space (MSP, GEN), logit space (Energy, Maxlogit), feature space (Mahalanobis, KNN), penultimate activation manipulations (ReAct, ASH), to a combination of logit and feature space (VIM). 
We employ the SOTA Agree-to-Disagree algorithm (A2D) \cite{dong2024multiood} and the combination of A2D and NP-Mix algorithm (AN) \cite{dong2024multiood} as the baseline methods. 


\noindent \textbf{Implementation and Hardware.}
We use the SlowFast model \cite{feichtenhofer2019slowfast} as our feature extractor, using pre-trained weights from Kinetics-600 \cite{kay2017kinetics}. All parameters of the SlowFast model are frozen throughout the training process to maintain consistency. All models are implemented on the MultiOOD codebase \citep{dong2024multiood} and run on a multi-NVIDIA RTX 6000 Ada workstations.
See full details in Appx.~\ref{appx:implementation}.


\subsection{Comparison with State-of-the-Art (SOTA)}
\label{subsec:comparisonwsota}
\vspace{-0.05in}
Tables~\ref{TableMain} and \ref{TablefarHMDB} and Appx. Tables~\ref{TableFarKinetics} and ~\ref{Tablethreemodal} show the effectiveness of \method in comparison to SOTA methods on 2 tasks and 9 OOD detection methods with all 5 datasets.
\method consistently achieves the best or near-best values for key metrics, including FPR95, AUROC, and ID Accuracy. 

\begin{figure}[t]
\includegraphics[width=0.47\textwidth]{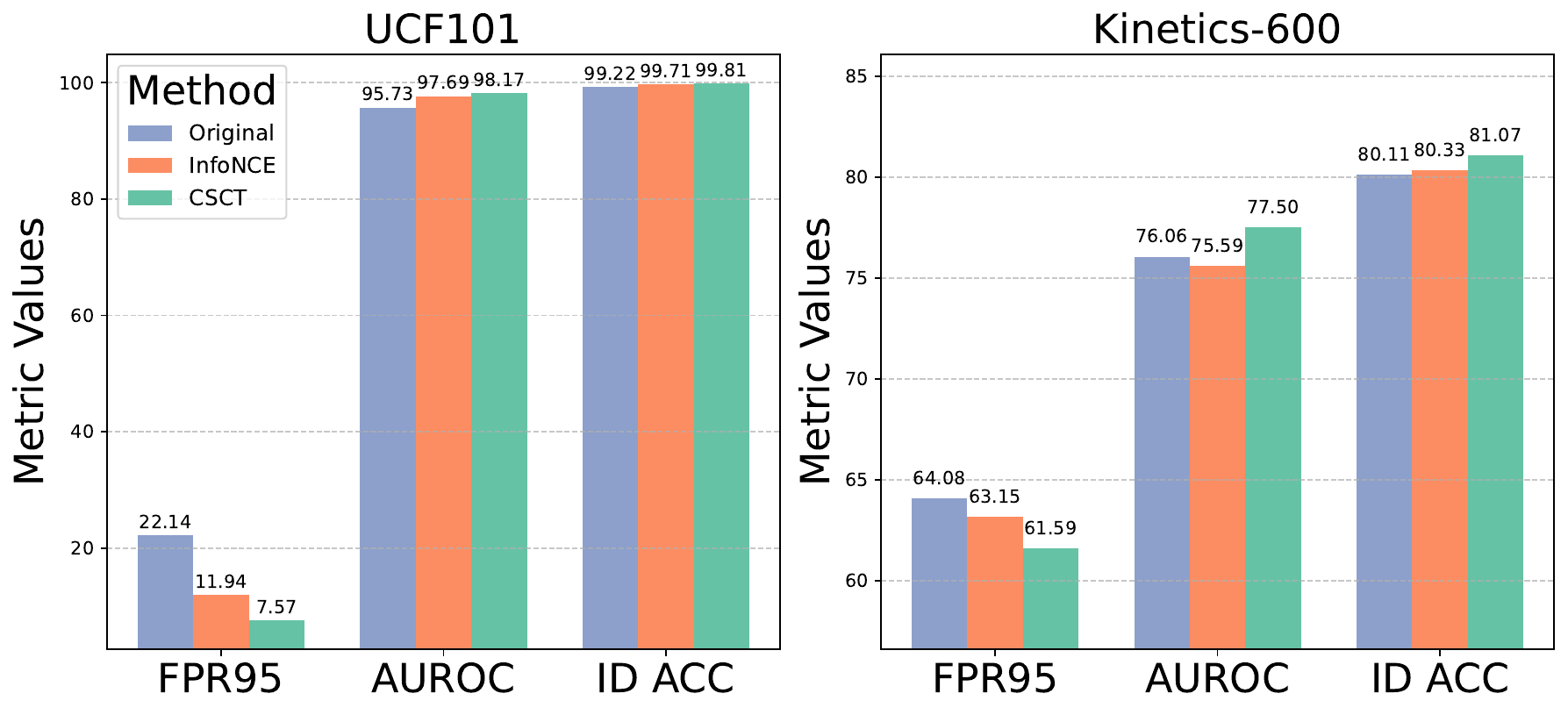} 
\vspace{-0.15in}
   \caption{Ablation study on contrastive learning methods. The experiment is conducted using MSP on Near-OOD detection with UCF101, and  Kinetics-600 dataset. Original: plain MSP. InfoNCE: the classic contrastive learning method \cite{oord2018representation}. CSCT:  proposed Cohesive-Separate Contrastive Training.} 
   \label{Fig:ablation1}
   \vspace{-0.25in}
\end{figure}

\noindent \textbf{\method is model-agnostic for diverse OOD detection methods.}
We evaluate \method across nine base OOD methods with different training strategies, including probability space, logit space, feature space, penultimate activation manipulations, and combinations of logit and feature space techniques. 
Despite this diversity, \method consistently improves all these OOD methods on nearly every evaluation metric. 
For example, on the HMDB51 dataset with the Energy OOD detection, \method significantly reduces FPR95 from 43.36 to 35.07, while increasing AUROC from 87.46 to 89.52 and ID ACC from 89.32 to 92.16. 
These results highlight \method's ability to markedly lower false positive rates and boost identification accuracy for various OOD models.

\noindent \textbf{\method is effective on multiple OOD tasks.}
We conduct experiments on both Near (Tab.~\ref{TableMain}) and Far (Tab.~\ref{TablefarHMDB} and Tab.~\ref{TableFarKinetics}) OOD detection tasks, and \method consistently demonstrates superior performance across both. We observe that \method consistently outperforms the baseline methods on both Near-OOD and Far-OOD detection tasks, showing its applicability and effectiveness. 
As shown in Fig.~\ref{Fig:intro1} and Tab.~\ref{TableMain}, \method yields substantial performance improvements, with gains of around 10\% across all metrics for Near-OOD and up to 80\% in certain metrics for Far-OOD detection.

\noindent \textbf{\method performs efficiently across diverse datasets.}
We evaluate \method on a total of five datasets, each representing different video styles, ranging from digitized movies and daily YouTube videos to cartoon figures and kitchen environments. Across every OOD method and dataset combination, \method consistently lowers FPR95 values and enhances AUROC and ID ACC. This demonstrates its robust capacity to identify Near-OOD samples with higher accuracy and reliability. In datasets like HMDB51 and Kinetics600, \method achieves particularly notable improvements, especially in metrics such as AUROC, where it significantly surpasses other methods. This indicates \method's adaptability and stable performance across diverse datasets and OOD detection models, reinforcing its broad applicability and alignment with improved OOD detection outcomes.

More experimental results including Far-OOD detection with Kinetics-600 as the ID, and Near-OOD detection with full modality training are in Appx. \ref{appx:additional}.

\begin{figure}[t]
    \centering
    \begin{subfigure}{0.232\textwidth}  
        \centering
        \includegraphics[width=\linewidth]{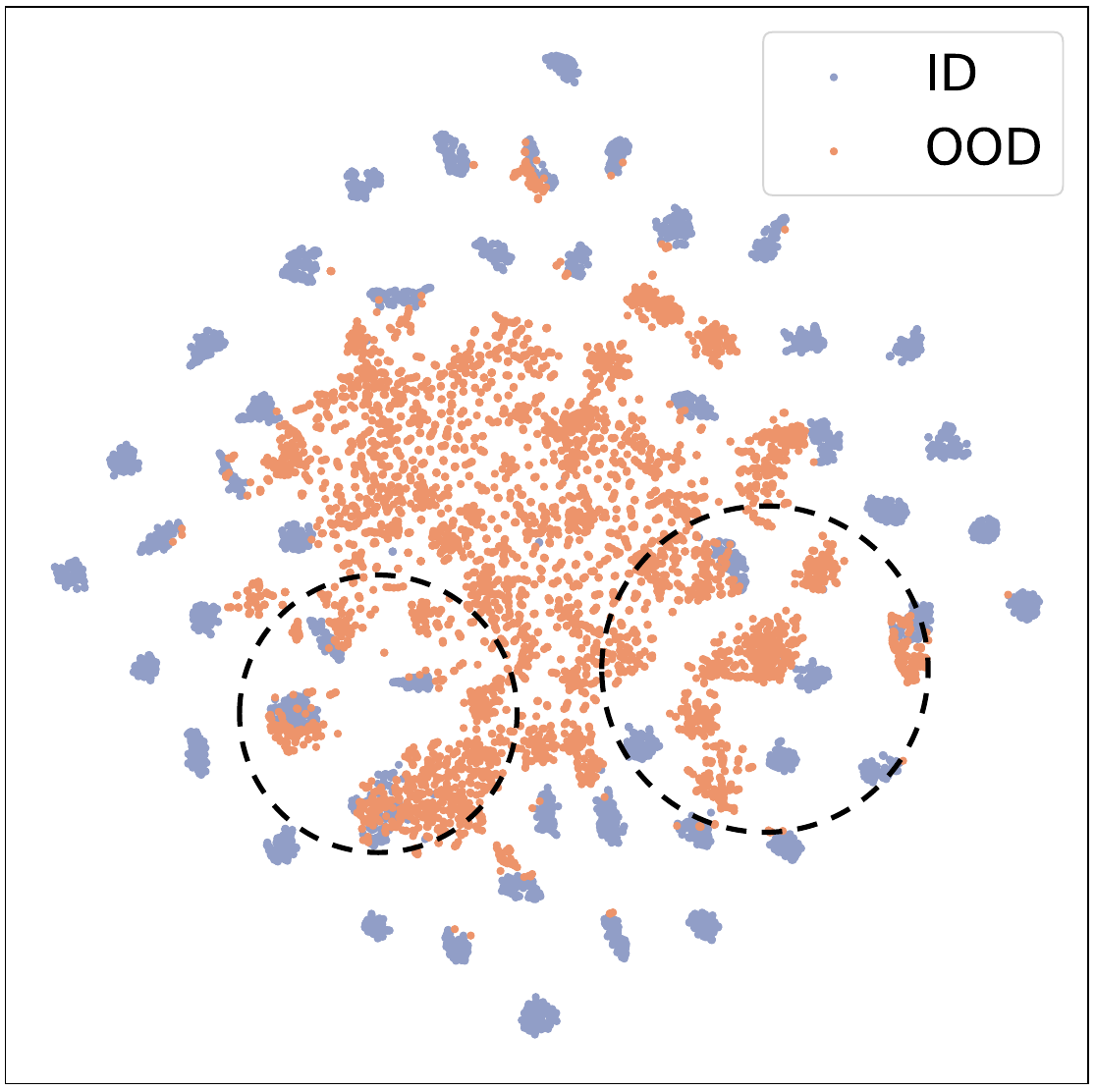} 
        \caption{Embeddings without \method.}  
        \label{Fig:sub1}  
    \end{subfigure}
    \hfill
    \begin{subfigure}{0.235\textwidth} 
        \centering
        \includegraphics[width=\linewidth]{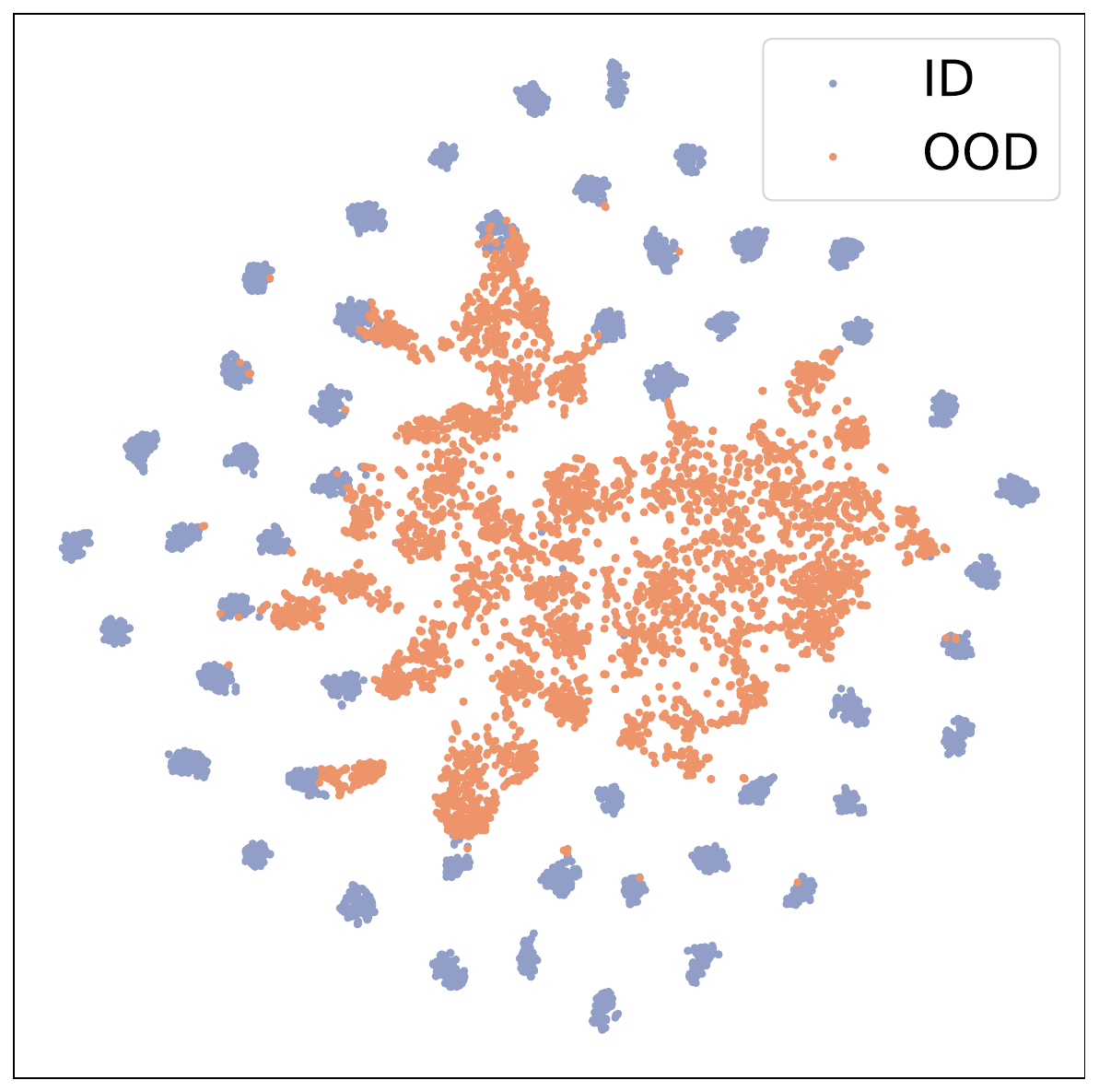}  
        \caption{\method-enhanced embeddings.}
        \label{Fig:sub2}
    \end{subfigure}
    \vspace{-0.1in}
    \caption{Visualization of the learned embeddings on ID and OOD data using t-SNE on the UCF101 50/51 dataset before and after training with \method. We observe better separation after using \method.
    } 
    \label{Fig:visualization}
\vspace{-1.7em}
\end{figure}

\noindent \textbf{Qualitative Analysis.} As shown in Fig.~\ref{Fig:visualization}, we provide visualizations of the learned embeddings for ID and OOD data using t-SNE on the UCF101 50/51 dataset before and after training with \method. Embeddings learned with \method exhibit greater robustness, with minimal overlap between ID and OOD embeddings, further showing its effectiveness in enhancing class separability and overall detection reliability.

\vspace{-0.1in}
\subsection{Ablation Studies}
\vspace{-0.1in}
We assess our method by employing different contrastive learning methods (\S \ref{subsec:contrastive}), and using fixed discrepancy intensification ratios instead of our adaptive one (\S \ref{subsec:proration}).

\noindent \textbf{Ablation Study on CSCT (Step 1).}
We assess the impact of various contrastive learning methods on near-OOD detection performance using MSP (see Fig.~\ref{Fig:ablation1}). The methods compared include the original MSP (no contrastive learning), InfoNCE (a widely-used contrastive method) \cite{oord2018representation}, and our \method using CSCT. 
Results show that CSCT consistently outperforms other approaches on UCF101 and Kinetics-600 datasets. Specifically, CSCT achieves the lowest FPR95 and highest AUROC, showing improved detection accuracy.
Also, CSCT demonstrates the highest ID ACC across datasets, highlighting its strength in maintaining intra-class cohesion and enhancing inter-class separation.

\noindent \textbf{Ablation Study on Pro-ratio Discrepancy Intensification.}
To investigate the impact of varying levels of discrepancy intensification, we substitute the adaptive $ \mathbf{I} $ with several fixed rates. As presented in Table~\ref{Tablefix-adaptive}, we replace $ \mathbf{I} $ with $ \left \{ 0.1, 0.3, 0.5, 0.7\right \} $ and report the results on the UCF101 and Kinetics-600 Near-OOD datasets. 
From the table, it is evident that any fixed, pre-defined rate fails to outperform \method. For example, while lower fixed rates (e.g., 0.1 or 0.3) demonstrate improved performance on UCF101, they perform poorly on Kinetics-600. Conversely, higher fixed rates lead to more balanced results between the two datasets but still cannot match the overall effectiveness of our adaptive method. This limitation arises because fixed rates do not accommodate the intra-class variances of ID samples, causing either under-emphasis or over-emphasis in intensifying discrepancies. By dynamically adjusting the discrepancy based on the data’s characteristics, our approach better balances intra-class cohesion and inter-class separability, enhancing the detection performance across datasets.
\vspace{-0.1in}


\begin{table}[]
\caption{Ablation study on the discrepancy intensification rate for multimodal Near-OOD using the Mahalanobis with UCF101 and Kinetics-600 datasets. It highlights that \method achieves the best performance across all metrics, showing the advantages of \method's adaptive intensification strategy over fixed intensification ratios.} 
\vspace{-0.1in}
\label{Tablefix-adaptive}
\centering
\setlength{\tabcolsep}{3pt} 
\fontsize{7}{10}\selectfont 
\begin{tabular}{c|ccc|ccc}
\toprule 
\multirow{2}{*}{Intensification Ratio} & \multicolumn{3}{c|}{UCF101} & \multicolumn{3}{c}{Kinetics-600} \\ \cline{2-7} 
                        & FPR95   & AUROC   & ID ACC  & FPR95     & AUROC    & ID ACC    \\ \midrule
0.1                & 10.19   & 97.88   & 99.51   & 92.49     & 50.05    & 80.52     \\
0.3                 & 6.89   & 98.44   & 99.81   & 93.43     & 50.58    & 79.91     \\
0.5                    & 7.86    & 97.99   & 99.71   & 93.59     & 50.10    & 79.93     \\ 
0.7                    & 8.25    & 98.28   & 99.81   & 92.68     & 50.81    & 80.42     \\ 
\method                    & \textbf{6.60}    & \textbf{98.68}   & \textbf{99.81}   & \textbf{89.27}     & \textbf{54.17}    & \textbf{81.07}     \\ \bottomrule 
\end{tabular}
\vspace{-0.2in}
\end{table}

%% file: tables/table2.tex
\begin{table*}[]
\caption{Multimodal Far-OOD Detection results using video and optical flow, with HMDB51 as the ID dataset ($\uparrow$ the higher the better; $\downarrow$ the lower the better). `AN' represents the combined use of Agree-to-Disagree and NP-Mix algorithms.}
\label{TablefarHMDB}
\vspace{-0.1in}
\centering
\renewcommand{\arraystretch}{1.00}
\setlength{\tabcolsep}{7pt} 
\fontsize{8}{10}\selectfont 
\begin{tabular}{r|cccccccc|c}
\toprule 
\multicolumn{1}{c|}{\multirow{3}{*}{Method}} & \multicolumn{8}{c|}{OOD Datasets}                                       & \multicolumn{1}{l}{\multirow{3}{*}{ID ACC}} \\ \cline{2-9}
\multicolumn{1}{c|}{}                        & \multicolumn{2}{c|}{Kinetics600}                                 & \multicolumn{2}{c|}{UCF101}                                     & \multicolumn{2}{c|}{HAC}                                        & \multicolumn{2}{c|}{EPIC-Kitchen}                      & \multicolumn{1}{l}{}                        \\ \cline{2-9}
\multicolumn{1}{c|}{}                        & \multicolumn{1}{l}{FPR95 $\downarrow$} & \multicolumn{1}{l|}{AUROC $\uparrow$}           & \multicolumn{1}{l}{FPR95 $\downarrow$} & \multicolumn{1}{l|}{AUROC $\uparrow$}          & \multicolumn{1}{l}{FPR95 $\downarrow$} & \multicolumn{1}{l|}{AUROC $\uparrow$}          & \multicolumn{1}{l}{FPR95 $\downarrow$} & \multicolumn{1}{l|}{AUROC $\uparrow$} & \multicolumn{1}{l}{}                        \\ \midrule
\multicolumn{1}{l|}{Energy}                  & 32.95                     & \multicolumn{1}{c|}{92.48}           & 44.93                     & \multicolumn{1}{c|}{87.95}          & 32.95                     & \multicolumn{1}{c|}{92.28}          & 8.10                      & 97.70                       & 87.23                                       \\
+AN                                          & 24.52                     & \multicolumn{1}{c|}{93.96}           & 36.49                     & \multicolumn{1}{c|}{89.67}          & 22.92                     & \multicolumn{1}{c|}{94.41}          & 6.96                      & 97.53                       & 86.89                                       \\
+Ours                                        & \textbf{21.89}            & \multicolumn{1}{c|}{\textbf{95.81}}  & \textbf{33.30}                     & \multicolumn{1}{c|}{\textbf{92.76}}          & \textbf{20.64}                     & \multicolumn{1}{c|}{\textbf{95.39}}          & \textbf{4.33}                      & \textbf{98.46}                       & \textbf{87.34}                              \\ \midrule
\multicolumn{1}{l|}{ASH}                     & 51.20                     & \multicolumn{1}{c|}{87.81}           & 53.93                     & \multicolumn{1}{c|}{84.22}          & 42.99                     & \multicolumn{1}{c|}{90.23}          & 19.95                      & 95.21                       & 86.20                                       \\
+AN                                          & 27.82                     & \multicolumn{1}{c|}{93.17}           & 38.43                     & \multicolumn{1}{c|}{89.52}          & 23.03                     & \multicolumn{1}{c|}{94.45}          & 6.84                      & 98.23                       & 86.20                                       \\
+Ours                                        & \textbf{26.45}            & \multicolumn{1}{c|}{86.62}           & \textbf{37.04}            & \multicolumn{1}{c|}{\textbf{89.95}} & \textbf{20.64}            & \multicolumn{1}{c|}{\textbf{95.39}} & \textbf{4.33}                      & \textbf{98.46}                       & \textbf{87.34}                              \\ \midrule
\multicolumn{1}{l|}{GEN}                     & 41.51                     & \multicolumn{1}{c|}{90.34}           & 46.18                     & \multicolumn{1}{c|}{87.91}          & 38.31                     & \multicolumn{1}{c|}{91.28}          & 8.21                      & 98.26                       & 87.23                                       \\
+AN                                          & 25.66                     & \multicolumn{1}{c|}{93.50}           & 37.40                     & \multicolumn{1}{c|}{91.19}          & 24.63                     & \multicolumn{1}{c|}{94.28}          & 5.25                      & 98.98                       & 86.89                                       \\
+Ours                                        & \textbf{7.18}             & \multicolumn{1}{c|}{\textbf{98.43}}  & \textbf{36.37}            & \multicolumn{1}{c|}{90.93}          & \textbf{21.43}            & \multicolumn{1}{c|}{\textbf{95.66}} & \textbf{3.19}                      & \textbf{99.30}                       & \textbf{87.34}                              \\ \midrule
\multicolumn{1}{l|}{KNN}                     & 22.69                     & \multicolumn{1}{c|}{95.01}           & 39.34                     & \multicolumn{1}{c|}{89.28}          & 20.75                     & \multicolumn{1}{c|}{96.02}          & 9.97                      & 97.92                       & 87.23                                       \\
+AN                                          & 15.05                     & \multicolumn{1}{c|}{96.96}           & 33.06                     & \multicolumn{1}{c|}{91.92}          & 13.45                     & \multicolumn{1}{c|}{97.25}          & 5.47                      & 98.97                       & 86.89                                       \\
+Ours                                        & \textbf{3.08}             & \multicolumn{1}{c|}{\textbf{99.50}}  & \textbf{31.81}            & \multicolumn{1}{c|}{\textbf{92.52}} & \textbf{14.48}            & \multicolumn{1}{c|}{\textbf{97.26}} & 5.87                      & \textbf{98.97}                       & \textbf{87.34}                              \\ \midrule
\multicolumn{1}{l|}{VIM}                     & 13.68                     & \multicolumn{1}{c|}{97.01}           & 33.87                     & \multicolumn{1}{c|}{91.45}          & 13.45                     & \multicolumn{1}{c|}{97.12}          & 5.93                      & 98.15                       & 87.23                                       \\
+AN                                          & 9.24                      & \multicolumn{1}{c|}{98.04}           & 26.45                     & \multicolumn{1}{c|}{92.34}          & 6.04                      & \multicolumn{1}{c|}{98.56}          & 5.36                      & 98.09                       & 86.89                                       \\
+Ours                                        & \textbf{0.01}             & \multicolumn{1}{c|}{\textbf{99.99}} & 27.59                     & \multicolumn{1}{c|}{90.86}          & \textbf{3.87}             & \multicolumn{1}{c|}{\textbf{99.16}} & \textbf{1.71}                      & \textbf{99.62}                       & \textbf{87.34}                              \\ \bottomrule 
\end{tabular}
\vspace{-0.15in}
\end{table*}

%% file: sec/5_Conlusion.tex
\section{Conclusion, Limitations, and Future Work}
\vspace{-0.1in}
The \method framework addresses the critical challenge of intra-class variability in multimodal OOD detection through adaptive, sample-specific prototype updates. Extensive experiments on diverse datasets and tasks demonstrate \method’s leading performance, achieving substantial improvements in OOD detection and, in some cases, perfect ID/OOD classification accuracy. These results highlight \method's potential for robust deployment in high-stakes applications.

\noindent \textbf{Limitations and Future Work.}
While \method performs well across a range of datasets, its effectiveness may vary with significantly larger or more complex multimodal datasets, which could introduce scalability challenges. 
Future work could focus on enhancing \method’s efficiency for these larger datasets and optimizing it for real-time applications, such as autonomous driving and healthcare, where prompt responses are essential. Additionally, exploring cross-domain OOD detection would broaden \method’s applicability, supporting its use in diverse real-world environments.
\vspace{-0.1in}

\section*{Broader Impact and Ethics Statement} 
\vspace{-0.1in}

\textbf{Broader Impact Statement}: \method significantly advances multimodal OOD detection by addressing intra-class variability, a crucial factor for reliable model performance in fields like healthcare, autonomous systems, and security. By enhancing model robustness against unknown data distributions, DPU empowers applications in dynamic, high-stakes environments where accuracy and reliability are paramount. This adaptability ensures systems remain effective when confronted with new or evolving data patterns.

\noindent
\textbf{Ethics Statement}: 
Our research complies with ethical standards, emphasizing privacy, fairness, and transparency. \method is designed to mitigate risks associated with biased predictions and potential privacy violations, particularly relevant in sensitive applications like surveillance and medical diagnosis. Promoting more consistent and fair OOD detection, \method helps reduce ethical concerns tied to unexpected model behaviors, especially meaningful for emerging ones. 


%% file: sec/X_suppl.tex
\setcounter{page}{1}
\maketitlesupplementary

\setcounter{table}{0}
\setcounter{figure}{0}

\renewcommand{\thetable}{\Alph{table}}
\renewcommand{\thefigure}{\Alph{figure}}

\section{Extended Related Work}
\label{appx:related}

\noindent \textbf{Out-of-Distribution Detection.}  
Out-of-Distribution (OOD) detection seeks to identify test samples diverging from the training distribution, while maintaining in-distribution (ID) classification accuracy. Common OOD methods include post hoc techniques and training-time regularization \cite{yang2022openood}. Post hoc approaches like Maximum Softmax Probability (MSP) \cite{oodbaseline17iclr}, enhanced by temperature scaling and input perturbation \cite{Liang_2018_ECCV}, compute OOD scores from model outputs, with improvements by methods like MaxLogit \cite{hendrycks2019anomalyseg} and energy-based approaches \cite{energyood20nips}. Activation-modifying techniques such as ReAct \cite{sun2021tone} and ASH \cite{djurisic2022extremely}, alongside distance-based approaches like Mahalanobis \cite{mahananobis18nips} and k-Nearest Neighbor (kNN) \cite{sun2022knnood}, leverage feature distances for detection. Recent hybrid methods, including VIM \cite{haoqi2022vim} and Generalized Entropy (GEN) \cite{liu2023gen}, integrate features and logits to improve scoring. Our method uniquely addresses intra-class variability within ID data, using dynamic prototype updates based on sample-specific variance, which enhances both ID and OOD detection.

\noindent
\textbf{Multimodal Learning.}
The multimodal learning field is progressing swiftly, driven by advancements in foundational models, larger datasets, and enhanced computational resources. Large pre-trained models, such as CLIP \cite{clip}, have substantially boosted performance across diverse multimodal tasks. However, scaling alone does not resolve core challenges, such as out-of-distribution issues \cite{10.1145/3701733} and model biases \cite{Wang_2023_ICCV}. Addressing edge cases and complex data formats in real-world scenarios remains essential for the safe and responsible deployment of these models.

\noindent \textbf{Multimodal Out-of-Distribution Detection.}
Recent research has expanded OOD detection to multimodal models, especially in vision-language systems \cite{ming2022delving,wang2023clipn}. Approaches like Maximum Concept Matching (MCM) \cite{ming2022delving} align visual features with textual concepts to define OOD scores. Meanwhile, CLIPN \cite{wang2023clipn} enhances CLIP by using contrasting prompts to distinguish between ID and OOD samples. However, these methods mainly focus on image-based benchmarks, limiting their use of complementary information from various modalities. In real-world applications, modalities such as LiDAR and cameras in autonomous driving or video, audio, and optical flow in action recognition provide valuable context that current methods do not exploit \cite{simonyan2014two}. Our approach introduces a dynamic multimodal framework that effectively leverages complementary modalities by adjusting prototype updates based on intra-class variability.

\section{Details on the Proposed \method}

\subsection{Robust Marginal Contrastive Learning}
\label{appx:robust-margin}
As briefly discussed in \S \ref{subsec:contrastive}, we consider a batch of video samples \( V_1, V_2, \ldots, V_n \) and their corresponding data from other modalities, \( M_1^i, M_2^i, \ldots, M_n^i \), where \( i \in \{ \text{flow, audio} \} \) represents the modality type, and \( n \) is the batch size. For each video sample \( V_j \), we define a positive set \( PS_j \), consisting of samples from the same class, and a negative set \( NG_j \), composed of samples from different classes. 
We start with the standard InfoNCE loss, a popular contrastive learning objective \cite{oord2018representation}, which is defined as:

\begin{equation}
\label{appx:eq:rmcl}
    \mathcal{L}_{rmcl} = \sum_{j=1}^{n} -\log \frac{f_{\text{pos}}}{f_{\text{pos}} + f_{\text{neg}}},
\end{equation}

where \( f_{\text{pos}} \) denotes the similarity scores for samples in the positive set \( PS_j \), and \( f_{\text{neg}} \) represents the similarity scores for samples in the negative set \( NG_j \).

To further improve robustness and sensitivity to inter-class distinctions, we introduce an angular margin \( m \) for positive samples to further enhance robustness by emphasizing inter-class distinctions, redefining \( f_{\text{pos}} \) as:

\begin{equation}
    f_{\text{pos}} = \sum_{b \in PS_j} \exp\left(\frac{\cos\left(\theta_{b,j} + m\right)}{t}\right),
\end{equation}

where \( t \) is a temperature hyper-parameter, and \( \theta_{b,j} \) denotes the arc-cosine similarity between representations \( F_b^i \) and \( F_j^i \) within the same modality. For the negative set \( NG_j \), we similarly define:

\begin{equation}
    f_{\text{neg}} = \sum_{b \in NG_j} \exp\left(\frac{\cos\left(\theta_{b,j}\right)}{t}\right).
\end{equation}

This contrastive objective encourages the model to maximize intra-class similarity (positive pairs) and minimize inter-class similarity (negative pairs).

\subsection{Final OOD Detection and Integration}
\label{appx:detection}
As briefly discussed in \S \ref{subsec:proration}, our multimodal OOD model makes predictions by leveraging: (\textit{i}) the joint probability distribution inferred across all modalities and (ii) separate predictions from each modality independently.

The model employs \( M \) feature extractors \( g_i(\cdot) \) for each modality, alongside a shared classifier \( h(\cdot) \). Each feature extractor \( g_i(\cdot) \) generates an embedding \( F_i \) for its respective modality \( i \), which the classifier \( h(\cdot) \) then combines to produce a joint probability distribution \( \hat{p} \):
\begin{equation}
    \hat{p} = \text{softmax} \left( h \left( \left[ g_1(x_{i}^1), g_2(x_{i}^2), \dots, g_M(x_{i}^M) \right] \right) \right).
\end{equation}
This combined output provides an overall prediction for identifying whether \( x_i \) belongs to the ID or OOD classes.

Additionally, each modality \( k \) has its own classifier \( h_k(\cdot) \), which produces individual prediction probabilities \( \hat{p}^k = \text{softmax} \left( h_k \left( g_k \left( x^k \right) \right) \right) \). This modality-specific information complements the joint prediction, offering finer-grained insights into each modality's confidence in classifying a sample as ID or OOD.
The core OOD detection loss for a sample \( (x_i, y_i) \) is:
\begin{equation}
    {\mathcal{L}}_{base} = \sum_{k=1}^{M} CE \left( \hat{p}_i^k, y_i \right) + CE \left( \hat{p}, y_i \right),
\end{equation}
where \( CE(\cdot) \) is the cross-entropy loss, reinforcing accurate predictions across joint and individual modality outputs.

Finally, the complete objective function for OOD detection integrates additional regularization components from the previous steps:
\begin{equation}
    \mathcal{L} = {\mathcal{L}}_{base} + \delta \cdot {\mathcal{L}}_{csct} + {\mathcal{L}}_{pdi} + \kappa \cdot {\mathcal{L}}_{aos},
\end{equation}
where \( \delta \) and \( \kappa \) are HPs that balance the influence of the CSCT, discrepancy intensification, and adaptive outlier synthesis losses, thereby optimizing the model’s performance on OOD detection.

\subsection{Hyperparameter Optimization in \method}
\label{appx:hyperparameter}
As briefly discussed in \S \ref{ssec:task-impl-detail}, for the HPs \( \lambda \), \( \delta \), and \( \kappa \), we set their values to 2, 0.2, and 0.5, respectively, to ensure they are within a similar order of magnitude. The value of \( \beta \) is set to 0.8 to control the update speed, with smaller values leading to faster updates. The parameter \( \gamma \) is assigned a very small value to prevent division by zero.
The most influential HP is \( \mu \), which directly impacts the discrepancy intensification. Based on the fixed ratio proposed by \citet{dong2024multiood}, we aim to scale \( \mu \) to allow sufficient room for dynamic adjustment. Specifically, we multiply the base fixed ratio by \( \left\{ 2, 3, 4 \right\} \) to determine the final value of \( \mu \). A grid search process is employed to find the optimal value.
Regarding batch size, we generally observe that a larger batch size during training yields better performance, which is characteristic of contrastive learning. However, an exception is the EPIC-Kitchen dataset, where the best performance is achieved with a batch size of 16. We attribute this to the specific data quality and distribution within the EPIC-Kitchen dataset.


\input{./tables/table3}

\section{Additional Experimental Settings and Results}
\subsection{Datasets}
\label{appx:datasets}
As briefly discussed in \S \ref{subsec:data-eva}, we evaluate our method across five datasets: HMDB51 \cite{kuehne2011hmdb}, UCF101 \cite{soomro2012ucf101}, Kinetics-600 \cite{kay2017kinetics}, HAC \cite{dong2023SimMMDG}, and EPIC-Kitchens \cite{Damen2018EPICKITCHENS}.

\noindent
1) \textbf{HMDB51} \cite{kuehne2011hmdb} is a video action recognition dataset containing 6,766 video clips across 51 action categories. The clips are sourced from various media, including digitized movies and YouTube videos, and include both video and optical flow modalities.

\noindent
2) \textbf{UCF101} \cite{soomro2012ucf101} is a diverse video action recognition dataset collected from YouTube, containing 13,320 clips representing 101 actions. This dataset includes variations in camera motion, object appearance, scale, pose, viewpoint, and background conditions. It provides video and optical flow modalities.

\noindent
3) \textbf{Kinetics-600} \cite{kay2017kinetics} is a large-scale action recognition dataset with approximately 480,000 video clips across 600 action categories. Each clip is a 10-second snippet of an annotated action moment sourced from YouTube. Following \cite{dong2024multiood}, we selected a subset of 229 classes from Kinetics-600 to avoid potential overlaps with other datasets, resulting in 57,205 video clips. Video and audio modalities are available, with optical flow extracted at 24 frames per second using the TV-L1 algorithm \cite{zach2007duality}, yielding 114,410 optical flow samples.

\noindent
4) \textbf{HAC} \cite{dong2023SimMMDG} includes seven actions—such as `sleeping', `watching TV', `eating', and `running'—performed by humans, animals, and cartoon characters, with 3,381 total video clips. The dataset provides video, optical flow, and audio modalities.

\noindent
5) \textbf{EPIC-Kitchens} \cite{Damen2018EPICKITCHENS} is a large-scale egocentric video dataset collected from 32 participants in their kitchens as they captured routine activities. For our experiments, we use a subset from the Multimodal Domain Adaptation paper \cite{munro20multi}, which contains 4,871 video clips across the eight most common actions in participant P22's sequence (`put,' `take,' `open,' `close,' `wash,' `cut,' `mix,' and `pour'). The available modalities include video, optical flow, and audio.

\subsection{Tasks}
\label{appx:tasks}
As briefly discussed in \S \ref{ssec:task-impl-detail}, we assess our approach on two tasks: Near-OOD detection and Far-OOD detection \cite{dong2024multiood}.

For Near-OOD detection, we evaluate using four datasets. In EPIC-Kitchens 4/4, a subset of the EPIC-Kitchens dataset is divided into four classes for training as ID and four classes for testing as OOD, totaling 4,871 video clips. HMDB51 25/26 and UCF101 50/51 are similarly derived from HMDB51 \cite{kuehne2011hmdb} and UCF101 \cite{soomro2012ucf101}, containing 6,766 and 13,320 video clips, respectively. For Kinetics-600 129/100, a subset of 229 classes is selected from Kinetics-600 \cite{kay2017kinetics}, with approximately 250 clips per class, totaling 57,205 clips. In this setup, 129 classes are used for training (ID) and the remaining 100 for testing (OOD). We present the results of \{video, optical flow\} on all four datasets, and the results of \{video, optical flow, audio\} on Kinetics-600 dataset.

In the Far-OOD detection setup, either HMDB51 or Kinetics-600 is used as the ID dataset, with the other datasets serving as OOD datasets:

\noindent
\textbf{HMDB51 as ID}: We designate UCF101, EPIC-Kitchens, HAC, and Kinetics-600 as OOD datasets. Samples overlapping with HMDB51 are excluded from each OOD dataset to maintain distinct ID/OOD classes. For instance, 31 classes overlapping with HMDB51 are removed from UCF101, leaving 70 OOD classes, and 8 overlapping classes are removed from EPIC-Kitchens and HAC.

\noindent
\textbf{Kinetics-600 as ID}: We designate UCF101, EPIC-Kitchens, HAC, and HMDB51 as OOD datasets, excluding any ID class overlap with Kinetics-600. For example, 11 overlapping classes are removed from UCF101, leaving 90 OOD classes, while the original classes in EPIC-Kitchens, HAC, and HMDB51 are preserved as they have no overlap with Kinetics-600.

\subsection{Implementation Details}
\label{appx:implementation}
As briefly discussed in \S \ref{ssec:task-impl-detail}, we employ Slowfast \cite{feichtenhofer2019slowfast} as the feature extraction model (pre-trained weights from Kinetics-600 \cite{kay2017kinetics}), and all its parameters are frozen during the training process.
For the Cohesive-Separate Contrastive Training process, the decision margin \textit{m} is set to 10 degrees, the temperature hyper-parameter t is set to 0.05, and the $ \lambda $ is set to 2. For the Dynamic Prototype Approximation process, $ \beta $ is set to 0.8, $ \gamma $ is set to 1e-6, $ \eta \in Beta(10, 10)$,  and $ \mu $ is set to 1.0, 1.2 , 0.2, 3.2 for HMDB, UCF, Kinetics and EPIC-Kitchen, respectively. For Pro-ratio Discrepancy Intensification, we activate it after 2 epoches of dynamic prototype approximation. We use fixed intensification rate in the first two epoches. For the final loss function, $ \delta $ is set to 0.2, and $ \kappa $ is set to 0.5. We use an AdamW optimizer with a learning rate 1e-4 and a batch size of 64 (for Near-OOD detection on EPIC-Kitchen dataset, the batch size is 16).

\input{./tables/table4}

\subsection{Case Study on Discrepancy}
\label{appx:intro-case}

\begin{figure}[t]
\begin{center}
\includegraphics[width=\linewidth]{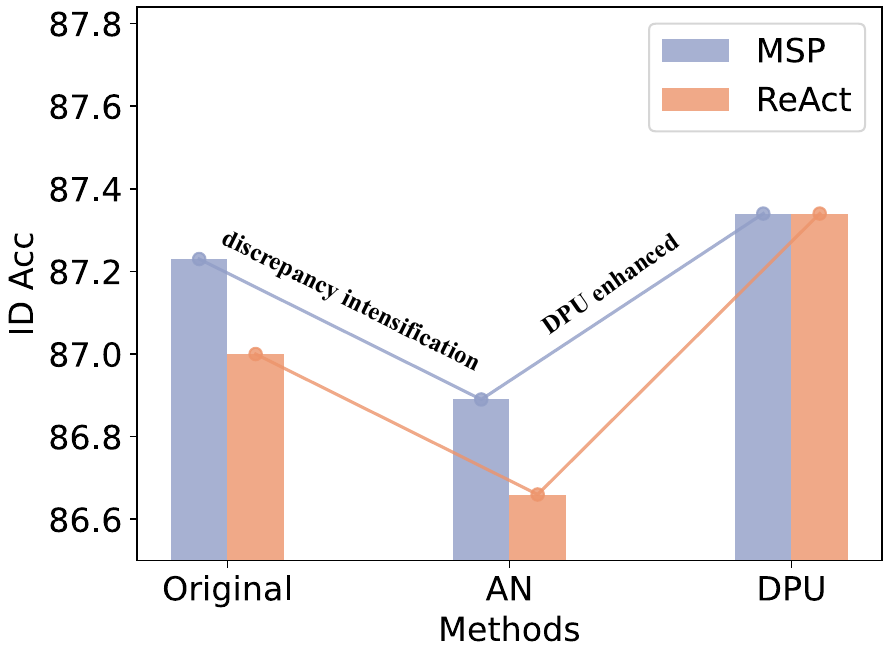} 
   \caption{The ID accuracy declines after using uniform discrepancy intensification in the SOTA framework \cite{dong2024multiood} (denoted as `AN'; the middle bars), and the accuracy improves using our proposed \method (the right bars). 
   This figure presents the results of MSP and ReAct in Far-OOD detection using HMDB51 as the ID dataset. 
   }
\label{Fig:intro2}
\end{center}
\end{figure}

As shown in Fig.~\ref{Fig:intro2}, we compare ID accuracy before and after applying both uniform discrepancy intensification \cite{dong2024multiood} and our \method. The results demonstrate a significant decline in ID accuracy with uniform discrepancy intensification, while \method enhances accuracy beyond the original levels. This improvement occurs because uniform discrepancy intensification, when applied to samples near the class center—which typically exhibit consistent predictions across modalities—disrupts this consistency, leading to degraded performance. In contrast, \method adaptively adjusts the degree of discrepancy intensification on a sample-wise basis, avoiding excessive intensification on class-center samples and thereby achieving new state-of-the-art ID accuracy.

\subsection{Additional Results}
\label{appx:additional}
As previously discussed in \S \ref{subsec:comparisonwsota}, we report additional experimental results here. Tab.~\ref{TableFarKinetics} reports the results of \method for Far-OOD detection with Kinetics-600 as the in-distribution (ID) dataset. The results reveal consistent performance gains over five baseline methods across four diverse OOD datasets. For example, on the UCF101 dataset \cite{soomro2012ucf101}, \method achieves a substantial 76.16\% reduction in FPR95 and a 32.82\% increase in AUROC when used in conjunction with VIM \cite{wang2022ofa}. These improvements underscore \method’s robustness in identifying far-OOD samples effectively.
Tab.~\ref{Tablethreemodal} presents the results of \method for Near-OOD detection on Kinetics-600, using three input modalities: video, optical flow, and audio. Here, \method again demonstrates superior performance compared to baseline methods, underscoring its adaptability and effectiveness across multiple modalities.

%% file: tables/table3.tex
\begin{table*}[]
\caption{Multimodal Far-OOD Detection using video and optical flow, with Kinetics-600 as ID. $\uparrow$ indicates larger values are better and vice versa. `AN' denotes employing Agree-to-Disagree and NP-Mix algorithm at the same time.}
\label{TableFarKinetics}
\centering
\setlength{\tabcolsep}{7pt} 
\fontsize{9}{10}\selectfont 
\begin{tabular}{r|cccccccc|c}

\toprule 
\multicolumn{1}{c|}{\multirow{3}{*}{Method}} & \multicolumn{8}{c|}{OOD Datasets}                                                                                                                                                                                                                             & \multicolumn{1}{l}{\multirow{3}{*}{ID ACC $\uparrow$}} \\ \cline{2-9}
\multicolumn{1}{c|}{}                        & \multicolumn{2}{c|}{HMDB51}                                 & \multicolumn{2}{c|}{UCF101}                                     & \multicolumn{2}{c|}{HAC}                                        & \multicolumn{2}{c|}{EPIC-Kitchen}                      & \multicolumn{1}{l}{}                        \\ \cline{2-9}
\multicolumn{1}{c|}{}                        & \multicolumn{1}{l}{FPR95 $\downarrow$} & \multicolumn{1}{l|}{AUROC $\uparrow$}           & \multicolumn{1}{l}{FPR95 $\downarrow$} & \multicolumn{1}{l|}{AUROC $\uparrow$}          & \multicolumn{1}{l}{FPR95 $\downarrow$} & \multicolumn{1}{l|}{AUROC $\uparrow$}          & \multicolumn{1}{l}{FPR95 $\downarrow$} & \multicolumn{1}{l|}{AUROC $\uparrow$} & \multicolumn{1}{l}{}                        \\ \midrule
\multicolumn{1}{l|}{Energy}                  & 72.64                     & \multicolumn{1}{c|}{71.75}           & 70.12                     & \multicolumn{1}{c|}{71.49}          & 61.50                     & \multicolumn{1}{c|}{74.99}          & 43.66                      & 82.05                       & 73.14                                       \\
+AN                                          & 63.27                     & \multicolumn{1}{c|}{74.17}           & 67.20                     & \multicolumn{1}{c|}{74.50}          & 56.69                     & \multicolumn{1}{c|}{80.20}          & 34.07                      & 87.49                       & 73.67                                       \\
+Ours                                        & \textbf{59.93}            & \multicolumn{1}{c|}{\textbf{79.47}}  & \textbf{45.53}                     & \multicolumn{1}{c|}{\textbf{84.46}}          & \textbf{47.41}                     & \multicolumn{1}{c|}{\textbf{82.99}}          & \textbf{31.59}                      & \textbf{89.24}                       & \textbf{76.74}                              \\ \midrule
\multicolumn{1}{l|}{ASH}                     & 71.62                     & \multicolumn{1}{c|}{76.66}           & 69.36                     & \multicolumn{1}{c|}{72.38}          & 47.85                     & \multicolumn{1}{c|}{83.49}          & 34.38                      & 88.05                       & 72.20                                       \\
+AN                                          & 63.65                     & \multicolumn{1}{c|}{79.14}           & 62.15                     & \multicolumn{1}{c|}{75.32}          & 46.47                     & \multicolumn{1}{c|}{85.21}          & 36.06                      & 88.69                       & 72.53                                       \\
+Ours                                        & \textbf{60.57}            & \multicolumn{1}{c|}{78.74}           & \textbf{45.52}            & \multicolumn{1}{c|}{\textbf{84.46}} & 47.41            & \multicolumn{1}{c|}{82.99} & \textbf{31.59}                      & \textbf{89.24}                       & \textbf{76.74}                              \\ \midrule
\multicolumn{1}{l|}{GEN}                     & 68.47                     & \multicolumn{1}{c|}{78.43}           & 64.80                     & \multicolumn{1}{c|}{73.94}          & 49.53                     & \multicolumn{1}{c|}{83.67}          & 36.81                      & 85.11                       & 73.14                                       \\
+AN                                          & 61.67                     & \multicolumn{1}{c|}{80.63}           & 57.03                     & \multicolumn{1}{c|}{77.97}          & 41.97                     & \multicolumn{1}{c|}{86.98}          & 40.09                      & 85.10                       & 73.67                                       \\
+Ours                                        & \textbf{53.92}             & \multicolumn{1}{c|}{\textbf{84.46}}  & \textbf{45.62}            & \multicolumn{1}{c|}{\textbf{85.56}}          & 49.05            & \multicolumn{1}{c|}{83.30} & \textbf{39.87}                      & \textbf{85.72}                       & \textbf{76.74}                              \\ \midrule
\multicolumn{1}{l|}{KNN}                     & 71.08                     & \multicolumn{1}{c|}{78.84}           & 68.62                    & \multicolumn{1}{c|}{74.33}          & 57.00                     & \multicolumn{1}{c|}{82.53}          & 41.83                      & 82.32                       & 73.14                                       \\
+AN                                          & 64.60                     & \multicolumn{1}{c|}{80.28}           & 67.44                     & \multicolumn{1}{c|}{77.95}          & 53.63                     & \multicolumn{1}{c|}{83.34}          & 37.65                      & 83.55                       & 73.67                                       \\
+Ours                                        & \textbf{61.33}             & \multicolumn{1}{c|}{\textbf{84.57}}  & \textbf{20.83}            & \multicolumn{1}{c|}{\textbf{95.85}} & \textbf{52.92}            & \multicolumn{1}{c|}{\textbf{83.47}} & 39.76                      & \textbf{85.37}                       & \textbf{76.74}                              \\ \midrule
\multicolumn{1}{l|}{VIM}                     & 72.25                     & \multicolumn{1}{c|}{71.18}           & 70.72                     & \multicolumn{1}{c|}{72.58}          & 59.48                     & \multicolumn{1}{c|}{75.46}          & 43.14                      & 82.69                       & 73.14                                      \\
+AN                                          & 63.25                      & \multicolumn{1}{c|}{73.96}           & 66.31                     & \multicolumn{1}{c|}{74.08}          & 53.82                      & \multicolumn{1}{c|}{81.06}          & 34.45                      & 87.67                       & 73.67                                       \\
+Ours                                        & \textbf{60.77}             & \multicolumn{1}{c|}{\textbf{78.92}} & \textbf{16.86}                     & \multicolumn{1}{c|}{\textbf{96.40}}          & \textbf{46.73}             & \multicolumn{1}{c|}{\textbf{83.50}} & \textbf{29.72}                      & \textbf{89.77}                       & \textbf{76.74}                              \\ \bottomrule 
\end{tabular}
\end{table*}

%% file: tables/table4.tex
\begin{table}[]
\caption{Multimodal Near-OOD Detection using video, optical flow, and audio. $\uparrow$ indicates larger values are better and vice versa. `AN' denotes employing Agree-to-Disagree and NP-Mix algorithm at the same time.}
\label{Tablethreemodal}
\centering
\setlength{\tabcolsep}{9pt} 
\fontsize{9}{10}\selectfont 
\begin{tabular}{r|ccc}
\toprule 
\multicolumn{1}{c|}{\multirow{2}{*}{Methods}} & \multicolumn{3}{c}{Kinetics-600 129/100}         \\ \cline{2-4} 
\multicolumn{1}{c|}{}                         & FPR95 $\downarrow$         & AUROC $\uparrow$         & ID ACC $\uparrow$        \\ \midrule
\multicolumn{1}{l|}{Energy}                   & 66.42          & 76.60          & 80.33          \\
+AN                                           & 63.81          & 77.89          & 80.82          \\
+Ours                                         & \textbf{62.64} & 77.57          & \textbf{82.21} \\ \midrule
\multicolumn{1}{l|}{ASH}                      & 63.48          & 78.11          & 79.54          \\
+AN                                           & 61.22          & 78.57          & 80.05          \\
+Ours                                         & 62.64          & \textbf{78.57} & \textbf{82.21} \\ \midrule
\multicolumn{1}{l|}{GEN}                      & 64.24          & 77.54          & 80.33          \\
+AN                                           & 63.55          & 77.92          & 80.82          \\
+Ours                                         & \textbf{60.95} & \textbf{78.50} & \textbf{82.21} \\ \midrule
\multicolumn{1}{l|}{VIM}                      & 66.38          & 76.59          & 80.33          \\
+AN                                           & 63.42          & 77.90          & 80.82          \\
+Ours                                         & \textbf{62.51} & \textbf{77.90} & \textbf{82.21} \\ \bottomrule 
\end{tabular}
   \vspace{-0.2cm}
\end{table}